\documentclass{article}

     \PassOptionsToPackage{numbers, compress}{natbib}
     \usepackage[final]{neurips_2020}

\usepackage[utf8]{inputenc} % allow utf-8 input
\usepackage[T1]{fontenc}    % use 8-bit T1 fonts
\usepackage{hyperref}       % hyperlinks
\usepackage{url}            % simple URL typesetting
\usepackage{booktabs}       % professional-quality tables
\usepackage{amsfonts}       % blackboard math symbols
\usepackage{nicefrac}       % compact symbols for 1/2, etc.
\usepackage{microtype}      % microtypography

\usepackage{graphicx}

\title{Improving Dialogue Breakdown Detection with Semi-Supervised Learning}

\author{%
  Nathan Ng\thanks{Work done at an internship at Google}, Marzyeh Ghassemi\\
  University of Toronto\\
  Vector Institute\\
  \texttt{\{nng, marzyeh\}@cs.toronto.edu}
  \And
  Narendran Thangarajan, Jiacheng Pan, Qi Guo\\
  Google Research\\
  \texttt{\{nart, jiachengpan, qiguo\}@google.com}
}

\begin{document}

\maketitle

\begin{abstract}
  Building user trust in dialogue agents requires smooth and consistent dialogue exchanges. However, agents can easily lose conversational context and generate irrelevant utterances. These situations are called dialogue breakdown, where agent utterances prevent users from continuing the conversation. Building systems to detect dialogue breakdown allows agents to recover appropriately or avoid breakdown entirely. In this paper we investigate the use of semi-supervised learning methods to improve dialogue breakdown detection, including continued pre-training on the Reddit dataset and a manifold-based data augmentation method. We demonstrate the effectiveness of these methods on the Dialogue Breakdown Detection Challenge (DBDC) English shared task. Our submissions to the 2020 DBDC5 shared task place first, beating baselines and other submissions by over 12\% accuracy. In ablations on DBDC4 data from 2019, our semi-supervised learning methods improve the performance of a baseline BERT model by 2\% accuracy. These methods are applicable generally to any dialogue task and provide a simple way to improve model performance.
\end{abstract}

\section{Introduction}
In recent years, voice assistants have become ubiquitous in the household, performing tasks for users through a conversational interface.
Given the informal nature of language in these settings, there are many ways in which the agent can misunderstand user commands, intent, and how to complete actions.
A vital part of ensuring that a user continues to interact with an agent is the user's confidence in the agent's ability to correctly and smoothly respond to their requests.
Relying on conversational rather than transactional dialogue is a core method of building this trust \citep{bickmore2001relational}. 

However, conversational dialogue is difficult to generate and can often lead to situations where the agent produces an utterance that the user is unable to properly respond to or that creates friction between the user and agent.
We refer to these situations as dialogue \textit{breakdown}, where the agent is prevented from completing the desired goal of the dialogue either by user exasperation or agent misunderstanding \citep{Martinovsky2003BreakdownIH}.
Detecting when breakdown occurs is an essential part of ensuring that the effects of the breakdown are mitigated, either by recovering when they occur or avoiding their creation altogether \citep{higashinaka-etal-2016-dialogue}.
As in other dialogue settings, gathering labelled data is difficult.
Data collection must either rely on interrupting user interactions or paying a third-party to rate dialogue after its completion, both of which are often intrusive, expensive, and affected by user bias \citep{doi:10.1287/isre.5.1.48}.
In these settings, semi-supervised learning methods are a natural way to fully utilize the limited labelled data by leveraging the large amounts of unlabelled data that are commonly available.

In this paper, we investigate two semi-supervised learning methods to improve performance on dialogue breakdown detection. 
We consider continued pre-training on the Reddit dataset \citep{baumgartner2020pushshift} as an approximation of the dialogue domain we would like to eventually fine-tune on.
We also consider self-supervised manifold based data augmentation (SSMBA) \citep{ng2020ssmba}, a data augmentation method that utilizes our further pre-trained model to generate new training examples. 

We evaluate these methods on the Dialogue Breakdown Detection Challenge (DBDC) \citep{higashinaka-etal-2016-dialogue} English shared task.
We submit our final models to the 2020 DBDC5 shared task, placing first in the English track. 
We beat baselines and other submissions by over 12\% accuracy, 0.135 F1 score, and 0.02 Jensen Shannon (JS) divergence.
In experiments on data from 2019 (DBDC4), our baseline model improves over previous challenge winners by over 13\% \citep{higashinaka2019overview}. 
The addition of our semi-supervised learning methods improves these baseline models further by 2\% accuracy,  0.02
F1 score, and 0.003 Jensen Shannon (JS) Divergence.
Although we specifically consider the task of dialogue breakdown detection, these semi-supervised techniques are applicable generally to any supervised dialogue task and provide a simple way to improve performance.

\section{Background and Related Work}
The Dialogue Breakdown Detection dataset and shared task \citep{higashinaka-etal-2016-dialogue} are both relatively new and under active development. 
The leading methods in prior competitions included context-enriched memory networks \citep{shim2019context}, BERT models with traditional dialogue features \citep{sugiyama2019dialogue}, LSTMs \citep{hendriksen2019lstm}, and ensembles of decision trees \citep{wang2019rsl19bd}. 
Other related datasets include the Alexa Topical-Chat dataset \citep{Gopalakrishnan2019} and the Google Schema-Guided Dialogue Dataset, which have been used for other tasks like intent prediction, slot filling, dialogue state tracking, and language generation, although none contain data for dialogue breakdown detection.
To our knowledge our work is the first that investigates semi-supervised learning techniques for this task.

Semi-supervised learning is a common method used to improve classifier performance in NLP \citep{devlin2018, dontstoppretraining2020}.
Currently, the dominant paradigm consists of two stages.
First a large language model is pre-trained on a general corpus of text.
The representations learned from this model are then used in downstream tasks by adding a classification layer and fine-tuning the model in a supervised fashion.
One such model is BERT \citep{devlin2018}, a large bidirectional transformer trained on a masked language modeling (MLM) objective.
BERT achieves strong performance across multiple tasks, making it our baseline model of choice.

Recent work has shown the value of continuing to pre-train the embeddings learned by BERT on a corpus of domain-specific text \citep{alsentzer-etal-2019-publicly, clinicalbert} to improve performance on downstream tasks in those domains.
Further pre-training on task-specific text \citep{dontstoppretraining2020} has also shown benefits.
Utilizing Reddit as a source of continued pre-training data has been investigated in claim detection tasks \citep{chakrabarty-etal-2019-imho}, but not for dialogue tasks.

Data augmentation in NLP is another semi-supervised learning method used to improve generalization and performance. 
Methods based on back-translation have been applied in the context of machine translation \citep{sennrich2016improving}, question answering \citep{wei2018fast}, and consistency training \citep{xie2019unsupervised}.
More recent work has used word embeddings \citep{wangyang2015thats} and LSTM language models \citep{fadaee2017data} to perform word replacement.
Self-supervised manifold based data augmentation (SSMBA) \citep{ng2020ssmba} is an augmentation method utilizing contextual language models to mask and reconstruct inputs and has been effective in multiple supervised NLP tasks. 

%We can frame BERT as a denoising autoencoder that learns to approximate the true data-generating manifold $P(x)$ \citep{bengio2013generalized}.
%By fine-tuning on top of the representations learned by such a model, our model's output class distribution must be smooth with respect to its manifold approximation.
%A core assumption of semi-supervised learning is the \textbf{manifold assumption}, that states that high dimensional data concentrates around a low-dimensional manifold \citep{chapelle2006semi}. 
%Methods that ensure smoothness in the class conditional distribution with respect to the underlying data manifold have been shown to improve performance and generalization in semi-supervised and self-supervised settings \citep{bachman2014learning,szegedy2014intriguing, sajjadi2016regularization}.

\section{Data}
Little public data exists for the task of dialogue breakdown detection, so 
we perform all experiments on the dataset distributed with the Dialogue Breakdown Detection Challenge (DBDC) \citep{higashinaka-etal-2016-dialogue, higashinaka2020overview} English shared task. 
This dataset consists of 411 English language dialogues between human users and conversational agents, each containing 10-20 utterances.
Since we predict breakdown only on agent dialogues, we have approximately 2,000 training examples, and a similar number of provided test examples. 
We set aside 200 of these training examples as a validation set.

\textbf{Breakdown Labels}
Each agent utterance is labelled by 15 third-party annotators with the labels No Breakdown (NB) indicating no breakdown in the conversation, Some Breakdown (SB) indicating noticeable breakdown in the dialogue, and Breakdown (B) indicating total breakdown in the dialogue.
Systems are tasked with predicting both the majority voted label as well as the matching the annotator label distribution.
The distribution of majority assigned labels in the training set is 56.8\% breakdown, 3.6\% some breakdown, and 39.6\% no breakdown.

\textbf{Pre-processing}
All data is preprocessed by tokenizing with a Wordpiece vocabulary of size 30k. 
We concatenate prediction utterances with one utterance of prior context, where the first sentence is the user context and the second sentence is the agent utterance to predict on.
This sentence pair forms the input examples to our models.
Augmented data is generated in a similar format.

\textbf{Evaluation Metrics}
We report the accuracy and F1 score of our model's prediction of the majority label.
Since we are also interested in matching the annotator label distribution, we report Jensen Shannon (JS) divergence.

\section{Experiments}

\subsection{Baseline Models}
Our baseline models are BERT\textsubscript{BASE} models with an MLP classifier head added to the classifier token. 
The input to our models consist of two sentences, the agent utterance we would like to predict breakdown on, and the prior human utterance as context. 
These two sentences are combined with the separator token between them as they would in a sentence pair classification task like natural language inference.
The model is then trained to minimize the KL divergence between the output label distribution and the annotator label distribution.
Compared to previous work on the DBDC shared task using BERT models \citep{sugiyama2019dialogue}, we use only the raw text features included in the dialogues.

\subsection{Continued Pre-training}
%Motivated by our framing of pre-training as manifold learning, we consider tasks like dialogue breakdown detection, where the data generating manifold is considerably different from the generic manifold learned by BERT. 
%We aim to adapt this generic manifold to the task domain by continued pre-training on a task domain corpus.
%Recent research has shown that this continued pre-training improves performance of downstream classifiers \citep{dontstoppretraining2020}.
We experiment on adapting BERT models pre-trained on generic English data to the dialogue domain by continued pre-training.
We train BERT\textsubscript{BASE} size models on the original tasks of masked language modeling and next sentence prediction.
For our continued pre-training experiments, we use the Reddit dataset \citep{baumgartner2020pushshift} as an approximation of the dialogue task domain.
This dataset consists of over 5 billion comments collected on Reddit between June 2005 and April 2019. 
To generate training pairs for BERT, we pair child comments with their corresponding parent comments similar to the way we pair prior user utterances and agent utterances during fine-tuning.
We perform ablations on the initialization, starting from a random initialization as well as BERT\textsubscript{BASE} initialization,
continuing to pre-train for 40 epochs.

\begin{figure}[t!]
\centering
\includegraphics[scale=0.33]{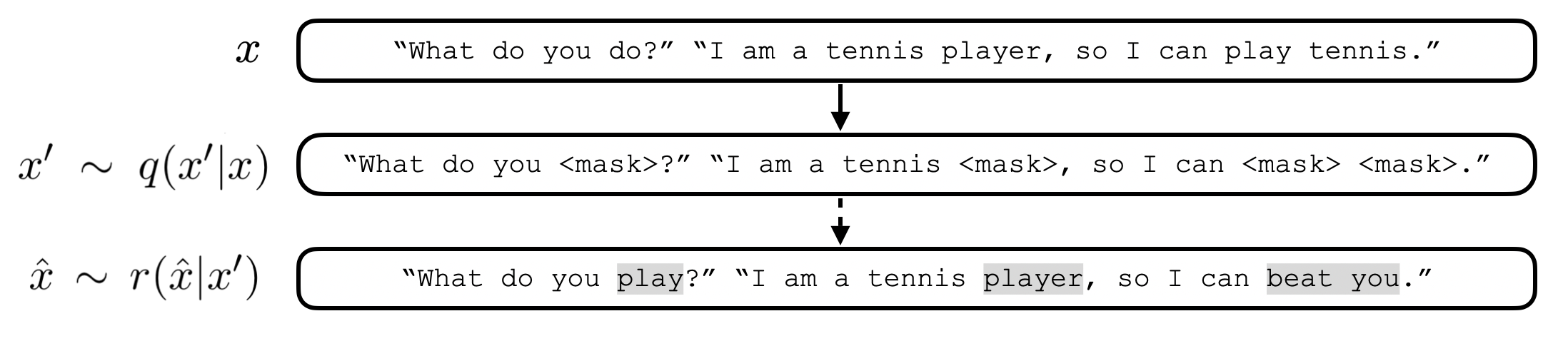}
\caption{To generate an augmented example with SSMBA, we apply the MLM corruption $q$ to the original sentence then reconstruct the corrupted sentence using our pre-trained BERT model $r$.}
\label{fig:ssmba_ex}
\end{figure}

\subsection{SSMBA Data Augmentation}
Self-Supervised Manifold-Based Data Augmentation (SSMBA) \citep{ng2020ssmba} is a data augmentation method that generates new examples in a manifold neighborhood of input examples by noising and reconstructing examples with denoising autoencoders (DAEs). 
We choose to investigate SSMBA since
compared to other more traditional rule based augmentation methods, SSMBA requires no task-specific knowledge and does not rely on class- or dataset-specific fine-tuning.
%However, it does require a DAE trained to approximate the data-generating manifold of the fine-tuning task. 
We adapt SSMBA to the dialogue setting by using our Reddit pre-trained model as our reconstruction model and applying noise to pairs of user-agent utterances. 
An example augmentation on an utterance pair is shown in Figure \ref{fig:ssmba_ex}.
Since augmented examples are not guaranteed to preserve the label of the original example, they are pseudo-labelled with a model trained on the original dataset.
Labelled examples are then combined with the original examples to form the final augmented dataset.
In our experiments, for each input training example, we apply a noising percentage of 20\% and generate 8 augmented examples.
Each augmented example is then soft pseudo-labelled.

\section{Results}

\begin{table}[t]
    \centering
    \begin{tabular}{lccc}
        \toprule
        System & Accuracy & F1 Score & JS Div\\
        \midrule
        DBDC4 BERT \citep{sugiyama2019dialogue} & 60.10 & - & 0.0580 \\
        \midrule
        Reddit Pre-training from Scratch & 66.12 & - & - \\
        Our BERT Baseline & 73.18 & 0.4938 & 0.0450 \\
        + Continued Reddit pre-training & 74.10 & 0.5040 & 0.0471 \\
        + SSMBA & 74.97 & \textbf{0.5170} & 0.0419 \\
        + Ensembling & \textbf{75.54} & 0.5128 & \textbf{0.0419} \\
        \bottomrule
    \end{tabular}
    \vspace{0.1cm}
    \caption{Accuracy, macro F1 score, and JS divergence of models on DBDC4 test data.}
    \label{tab:dbdc4}
\end{table}

\subsection{DBDC4}
Results and ablations on DBDC4 are presented in Table \ref{tab:dbdc4}.
DBDC4 baselines are BERT\textsubscript{BASE} models.
Ensembles are formed from the top 4 performing models.
Our baseline models already significantly outperform a comparable model \citep{sugiyama2019dialogue} from the DBDC4 competition, perhaps due to the simplification of the model architecture and eschewing of traditional dialogue features.

Pre-training only on Reddit data from a random initialization degrades accuracy below baseline levels. 
However, continued pre-training on Reddit starting from a BERT\textsubscript{BASE} initialization confers a gain of 1\% accuracy with similar F1 score and JS divergence. 
This difference in performance can be perhaps attributed to the structured English learned by BERT during the initial round of pre-training on Wikipedia and Books Corpus, which is absent in much of the more informal Reddit data.

The addition of SSMBA adds almost 1\% gain in accuracy, with a large drop in JS divergence.
Finally, ensembling confers another small gain in accuracy without any deterioration in JS divergence.
Although these gains are relatively small, they are noteworthy considering the small size of the original dataset.

\subsection{DBDC5}
Final results on DBDC5 are presented in Table \ref{tab:dbdc5}.
For DBDC5, F1 score is reported only on the breakdown label.
Shared task baselines are conditional random fields (CRF) \citep{higashinaka-etal-2016-dialogue}.
Our submission significantly outperforms the CRF baseline and the other submissions, beating the next best performing model by 12\% in accuracy, 0.135 in F1 score, and 0.02 in JS divergence.

\begin{table}[t]
    \centering
    \begin{tabular}{lccc}
        \toprule
        System & Accuracy & F1 Score (B) & JS Div\\
        \midrule
        CRF Baseline & 58.66 & 0.6928 & 0.3740 \\
        NUSLP & 61.75 & 0.6470 & 0.0907 \\
        \midrule
        Our Submission & \textbf{73.92} & \textbf{0.7823} & \textbf{0.0695} \\
        \bottomrule
    \end{tabular}
    \vspace{0.1cm}
    \caption{Accuracy, breakdown F1 score, and JS divergence of models on DBDC5 test data.}
    \label{tab:dbdc5}
\end{table}

\section{Conclusion}
Building trust and confidence in agents with conversational interfaces requires smooth dialogue that avoids breakdown.
Detecting dialogue breakdown, either before or while it happens, is an essential part in ensuring users have satisfactory  experiences with these agents.
We investigate two semi-supervised learning methods to leverage unlabelled data to improve breakdown detection, including continued pre-training on the Reddit dataset and SSMBA data augmentation.
We utilize these methods in our submission to the 5th Dialogue Breakdown Detection Challenge, beating other baselines and submissions by a large margin.
In ablations on previous test sets, we show that the addition of our semi-supervised methods improves our baseline models by over 2\% accuracy and reduces JS divergence by over 0.003.
These methods are simple and applicable to any dialogue task.
In the future we will continue to investigate applying these methods for intent prediction, slot filling, state tracking and language generation.

\bibliographystyle{plainnat}
\bibliography{neurips_2020}

\end{document}